\pdfoutput=1
\documentclass[sigconf, final, nonacm]{acmart}
\renewcommand\footnotetextcopyrightpermission[1]{} 

\AtBeginDocument{%
  \providecommand\BibTeX{{%
    \normalfont B\kern-0.5em{\scshape i\kern-0.25em b}\kern-0.8em\TeX}}}

\usepackage{times}
\usepackage{epsfig}
\usepackage{graphicx}
\usepackage{amsmath}

\usepackage{algorithm, algpseudocode}
\usepackage{multirow}
\usepackage{caption}
\usepackage{comment}
\usepackage{booktabs}
\usepackage{float}
\usepackage{xcolor}
\usepackage{siunitx}
\usepackage{indentfirst}
\usepackage{microtype}


\definecolor{citecolor}{RGB}{119,185,0}

\usepackage{pifont}
\usepackage{color}
\newcommand{\cmark}{\ding{51}}%
\newcommand{\xmark}{\ding{55}}%

\newlength\savewidth

\newcommand{\method}{RASR}
\newcommand{\task}{PairUAV}
\newcommand{\score}{0.003189}
\newcommand{\distonline}{0.003029}
\newcommand{\angleonline}{0.003350}

\title{RASR: Range-Aware Scale Recovery for Metric UAV Navigation}

\author{%
\texorpdfstring{%
Hongtao Liang$^{1}$ \quad Xinyu Shao$^{2,3}$ \quad Chenxu Wang$^{4}$ \quad Yiyao Wan$^{1}$\\
Jiahuan Ji$^{5}$ \quad Fangwei Ye$^{6}$ \quad Fuhui Zhou$^{5}$ \quad Qihui Wu$^{1}$}{Hongtao Liang et al.}}
\affiliation{%
  \institution{%
  $^{1}$College of Electronic and Information Engineering, Nanjing University of Aeronautics and Astronautics, China\\
  $^{2}$Shenzhen International Graduate School, Tsinghua University, China, $^{3}$Noah Ark Lab, Huawei, China\\
  $^{4}$College of Automation Engineering, Nanjing University of Aeronautics and Astronautics, China\\
  $^{5}$College of Artificial Intelligence, Nanjing University of Aeronautics and Astronautics, China\\
  $^{6}$College of Computer Science and Technology, Nanjing University of Aeronautics and Astronautics, China\\
  Emails: lianghongtao@nuaa.edu.cn, shaoxy23@mails.tsinghua.edu.cn, chenxuwang@nuaa.edu.cn, yiyaowan@nuaa.edu.cn, jiahuan.ji@nuaa.edu.cn, fangweiye@nuaa.edu.cn, zhoufuhui@ieee.org, wuqihui2014@sina.com}
  \city{\relax}
  \country{\relax}}

\renewcommand{\shortauthors}{Liang et al.}

\begin{document}
\flushbottom

\begin{abstract}
A central challenge in image-goal UAV navigation under Global Navigation Satellite System (GNSS) denial is estimating metric distance and heading between current and goal views. 
Dense pairwise geometry models capture relative scene structure, but without a calibrated metric scale, they cannot directly provide reliable distance estimates for navigation. Although global scale calibration corrects the dominant scale bias, the remaining errors vary systematically with distance.
In this paper, \underline{\textbf{R}}ange-\underline{\textbf{A}}ware \underline{\textbf{S}}cale \underline{\textbf{R}}ecovery (\textbf{\method{}}) is proposed, which complements global scale calibration with range-aware residual correction. RASR encodes pairwise geometry extracted by a frozen Matching And Stereo 3D Reconstruction (MASt3R) backbone as a compact descriptor and separates the scale-recovery core from task-specific command calibration. 
On the official online evaluation of the UAVs in Multimedia 2026 \task{} challenge, \method{} achieved a total error of \score{}, achieving a lower total error than global scale calibration alone. 
The results demonstrate that range-aware residual correction improves metric distance estimation beyond global scale calibration.
Code and materials are available at \href{https://github.com/lht-research/rasr-pairuav}{\textcolor{blue}{\url{https://github.com/lht-research/rasr-pairuav}}}.
\end{abstract}


\maketitle

\begin{figure*}[!t]
  \centering
  \includegraphics[width=0.92\textwidth]{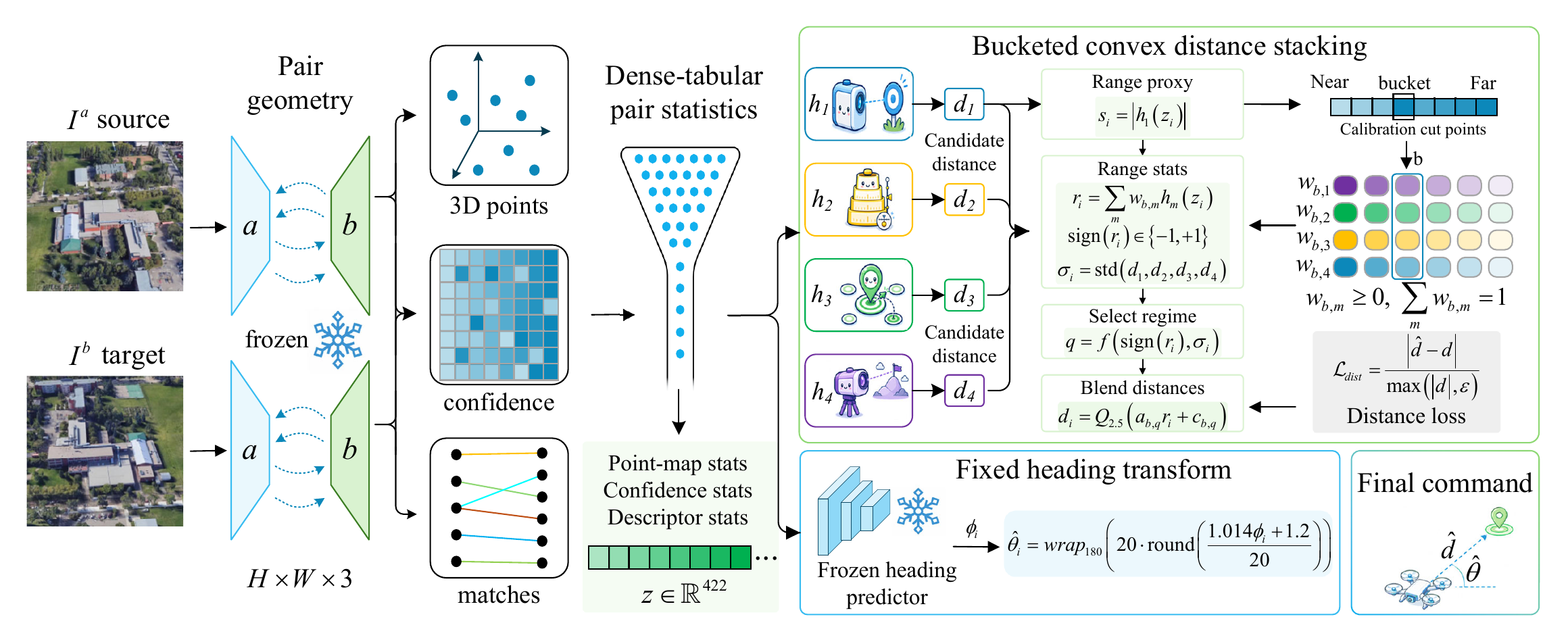}
  \caption{Overview of \method{}, where frozen pairwise geometry is summarized into a compact geometry descriptor, processed by the scale-recovery core, and passed to a separate protocol-specific calibration module.}
  \label{fig:pipeline}
\end{figure*}

\section{Introduction}

When Global Navigation Satellite System (GNSS) signals are unavailable, visual observations provide the goal-relative information needed for unmanned aerial vehicle (UAV) navigation.
Such goal-directed visual navigation is relevant to the final stages of autonomous landing, aerial delivery, and search-and-rescue missions~\cite{shah2023gnm,shah2023vint,sridhar2024nomad,deng2026anyimagenav,liang2026synergetic}.
In image-goal navigation, a reference image specifies the destination, while the UAV estimates the distance and heading to that destination from its current view.
The UAVs in Multimedia (UAVM) 2026 \task{} challenge formulates this task as metric distance and heading estimation from ordered image pairs~\cite{li2026lastmeterprecisionnavigationuavs,deuser2026UVA}.
Unlike retrieval-based localization, \task{} must estimate both metric distance and heading to form navigation commands.

Dense pairwise geometry models such as Dense and Unconstrained Stereo 3D Reconstruction (DUSt3R) and Matching And Stereo 3D Reconstruction (MASt3R) recover dense 3D structure from image pairs~\cite{wang2024dust3r,leroy2024mast3r}.
For aerial image pairs, this structure is informative for heading estimation, but its scale does not reliably correspond to physical distance.
A distance derived from this geometry therefore remains an uncalibrated proxy and must be mapped to metric units before it can support navigation commands.

Global scale calibration provides a straightforward baseline by fitting a single affine mapping from this proxy to metric distance.
It removes the dominant scale bias but applies the same mapping across the entire distance range.
On \task{}, we observe that the residuals after global calibration still vary systematically with range.
As in AbsRel-based monocular depth evaluation, relative distance error normalizes absolute error using the ground-truth distance with a small numerical floor.
A given absolute error is therefore penalized more heavily at short range~\cite{yin2023metric3d,piccinelli2024unidepth,yang2024depthanythingv2}.
A single global mapping alone cannot correct these range-dependent residuals.
Since ground-truth distance is unavailable at inference, the correction must instead be conditioned on a proxy predicted from each image pair.

To address this limitation, we propose Range-Aware Scale Recovery (\method{}), which converts frozen pairwise geometry into metric distance estimates.
\method{} compresses the geometry into a compact descriptor and uses four distance heads to produce candidate estimates that behave differently across distance ranges.
A predicted range proxy assigns each image pair to a range-proxy bucket, allowing fixed mixture weights to combine the candidate estimates differently across buckets.
A separate protocol-specific calibration module applies the final distance correction, heading calibration, and command-grid alignment required by \task{}.
On the scene-disjoint evaluation set, \method{} achieves lower relative distance error than global scale calibration when protocol-specific calibration is enabled for both evaluated variants.

\section{Related Work}

\textbf{Cross-view localization and metric commands.}
Cross-view localization usually recovers pose by retrieval against aerial or satellite references, with transformer matching, geometric layout correspondence, and hard negative sampling as common routes~\cite{zhu2022transgeo,zhang2023geodtr,deuser2023sample4geo}.
Recent work further studies orientation, view variation, and language-conditioned interfaces, allowing retrieval-style localization to cover more complex visual changes~\cite{hu2022beyondgeoloc,mi2024congeo,ye2025whereami}.
Retrieval-based formulations also dominate drone geolocalization.
University-1652 and Game4Loc provide representative benchmarks~\cite{zheng2020university1652,ji2025game4loc}, while Feature Segmentation and Region Alignment (FSRA) is a retrieval model for cross-view matching~\cite{dai2022fsra}.
Low-altitude self-positioning work instead targets urban metric localization~\cite{dai2024uavselfpositioning}.
These methods and benchmarks provide strong visual grounding, but a per-pair command interface additionally requires calibrated distance and heading from a single aerial image pair.

\textbf{Frozen pairwise geometry and scale.}
DUSt3R and MASt3R~\cite{wang2024dust3r,leroy2024mast3r} predict dense 3D structure from pairs or collections with broad transfer, and Visual Geometry Grounded Transformer (VGGT) and Fast3R~\cite{wang2025vggt,yang2025fast3r} extend this direction.
Later models add temporal consistency and persistent state~\cite{zhang2025monst3r,wang2025cut3r}, and Reloc3R strengthens camera-pose grounding~\cite{dong2025reloc3r}.
A parallel line shows that pretrained geometry becomes metric only once scale is calibrated, from Metric3D, Metric3D v2, and UniDepth~\cite{yin2023metric3d,hu2024metric3dv2,piccinelli2024unidepth} to affine-invariant depth models such as Depth Anything, Depth Anything V2, and Marigold~\cite{yang2024depthanything,yang2024depthanythingv2,ke2024marigold}.
This paper adopts the same geometry-first premise, while the distance-normalized objective motivates range-aware correction beyond a single global calibration.
These works commonly report AbsRel-style relative errors, whose distance-normalized penalty matches the distance term adopted here. Whether a range-aware estimator helps in that setting is left to future work.

\textbf{Image-goal navigation.}
Image-goal navigation studies how an embodied agent reaches a visual target rather than a semantic category~\cite{shah2023gnm,shah2023vint}.
NoMaD and AnyImageNav extend this setting toward general goal-conditioned navigation and any-view geometry~\cite{sridhar2024nomad,deng2026anyimagenav}.
Local feature matchers provide another related route: SuperGlue and LightGlue establish robust correspondences, but correspondence alone is not a metric displacement~\cite{sarlin2020superglue,lindenberger2023lightglue}.
Existing work therefore provides effective and complementary tools for place recognition, visual correspondence, and action learning, but does not directly recover calibrated distance and heading from a single aerial image pair.
This gap motivates our use of frozen pairwise geometry as the basis for navigation-ready metric estimation.

\section{\method{} Framework}

\method{} processes each ordered source--target image pair independently through a scale-recovery core and a separate protocol-specific calibration module.
As shown in Figure~\ref{fig:pipeline}, the core summarizes frozen pairwise geometry as a 422-dimensional dense-tabular pair-statistics vector that serves as the compact geometry descriptor.
Four distance heads produce candidate estimates from this vector.
The range proxy assigns each image pair to a routing bucket, whose convex weights combine the candidates into a stacked estimate.
The calibration module then uses the routing bucket, stacked estimate, and candidate spread to select the final distance correction.
A separate fixed transform then calibrates the frozen heading prediction.
All model weights and calibration parameters are fixed before hidden-test inference.

\subsection{Range-Aware Distance Formulation}

For pair $i$, let $x_i=(I_i^a,I_i^b)$ denote an ordered source--target image pair, where $I_i^a$ is the current view and $I_i^b$ is the goal view.
The corresponding ground-truth command is $y_i=(\theta_i,d_i)$, and the estimator returns $\hat y_i=(\hat\theta_i,\hat d_i)$.
We require
\begin{equation}
  \hat y_i = f(I_i^a,I_i^b;\Theta),
\end{equation}
where $\Theta$ is fixed before hidden-test inference. Each prediction therefore depends only on the current image pair and fixed model parameters.
This formulation excludes test-set graph optimization, global assignment, batch sorting, and corrections based on other hidden image pairs.

Frozen pairwise geometry captures relative structure, but distances inferred from it are not calibrated to the \task{} metric command space.
Let $u_i$ denote the distance proxy used by the global-calibration baseline.
This baseline applies one affine mapping, $\hat d_i=\alpha u_i+\beta$, across the full distance range.
\task{} evaluates distance using the relative error
\begin{equation}
  \ell_d(\hat d,d)=\frac{|\hat d-d|}{\max(|d|,\epsilon)},
  \label{eq:rel}
\end{equation}
where $\epsilon=0.5\,\mathrm{m}$ prevents numerical instability near zero distance.
Given a predictive distribution $p(d\mid x)$, the point estimate that minimizes the expected score solves
\begin{equation}
  \begin{aligned}
    c^\star(x)
    &=\arg\min_{c}\ \mathbb{E}_{d\sim p(\cdot\mid x)}
      \!\left[\frac{|c-d|}{\max(|d|,\epsilon)}\right] \\
    &=\arg\min_{c}\ \mathbb{E}\!\left[w(d)\,|c-d|\right],
  \end{aligned}
  \label{eq:wmed}
\end{equation}
where $w(d)=1/\max(|d|,\epsilon)$.
The minimizer of a weighted absolute deviation is the weighted median, so $c^\star$ is the median of $\tilde p(d)\propto p(d)/\max(|d|,\epsilon)$.
This weighting places greater emphasis on short distances.
Because the local predictive distribution can vary with distance, one global affine mapping cannot generally remove the residual pattern across the full range.
The appropriate correction therefore depends on the local predictive distribution, not only on a global scale factor.
Ground-truth distance is unavailable at inference and cannot directly select this correction.
For range-aware routing, \method{} instead derives a range proxy from the current image pair and uses it to condition residual correction.

\subsection{Scale Recovery Core}

For each ordered image pair, the scale-recovery core extracts pairwise geometry from the frozen MASt3R Vision Transformer Large/16 (ViT-L/16) checkpoint trained for metric reconstruction.
Each image is processed at a 512-pixel input resolution, with symmetric inference in both source-to-target and target-to-source directions.
The bidirectional outputs provide 3D point maps, cross-view point maps, confidence maps, and descriptor maps.
We summarize these outputs with 144 point-map statistics, 28 confidence statistics, and 250 descriptor statistics.
Concatenating these summaries yields a 422-dimensional dense-tabular pair-statistics vector $z_i$, which serves as the compact geometry descriptor and contains no scene identifiers or image metadata.

\begin{figure}[t]
  \centering
  \includegraphics[width=\linewidth]{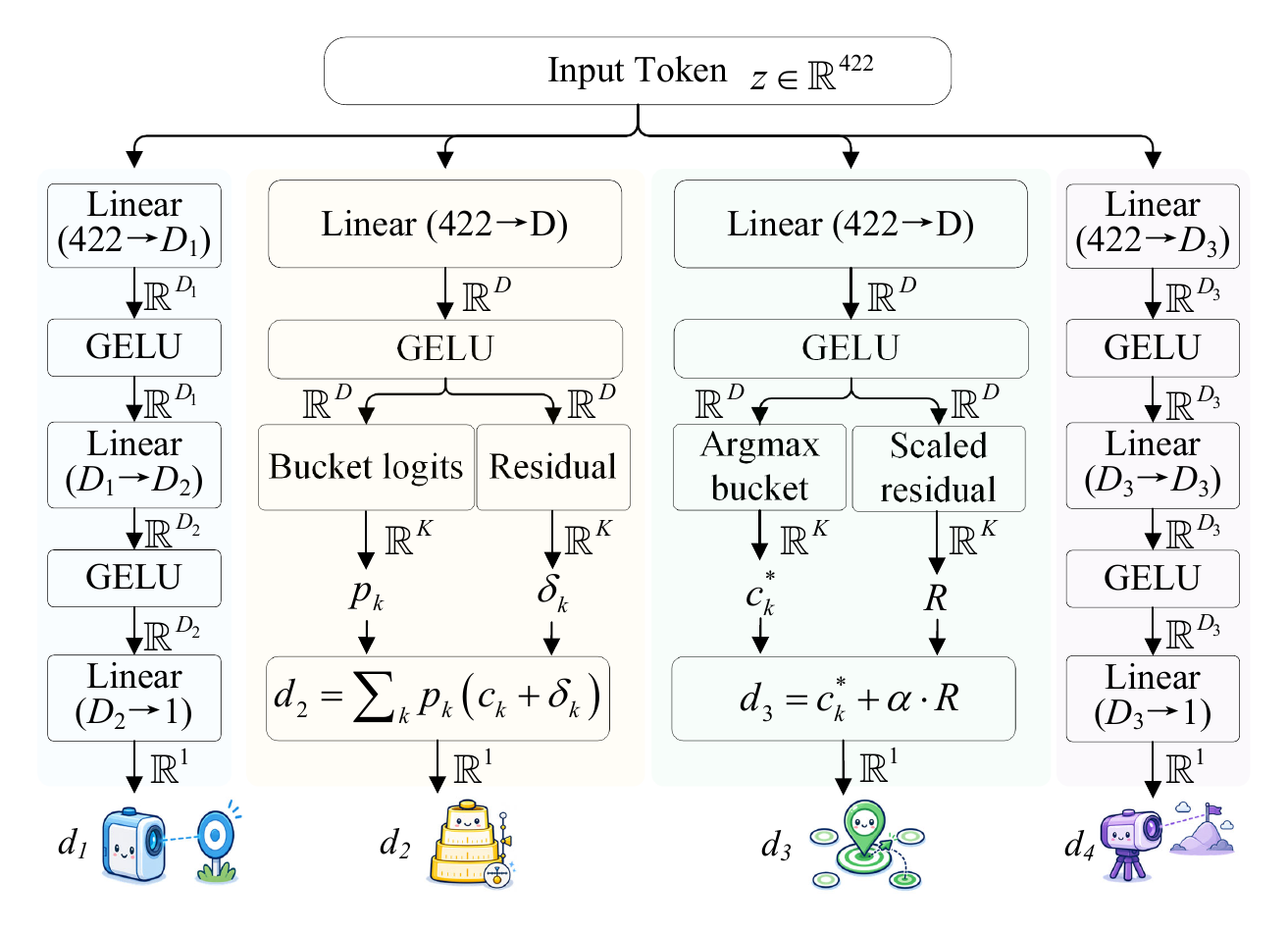}
  \caption{Four distance heads fitted on the calibration set and fixed at inference, producing candidate distances with complementary range behavior.}
  \label{fig:arch}
\end{figure}

The same input vector $z_i$ is passed to four distance heads.
For pair $i$, head $h_m$ produces a candidate distance $c_{i,m}=h_m(z_i)$; these candidates correspond to $d_1,\ldots,d_4$ in Figures~\ref{fig:pipeline} and~\ref{fig:arch}.
Figure~\ref{fig:arch} schematically summarizes the head designs used to generate the four candidates.
The released checkpoints cover three design families: hierarchical bucket-residual prediction, bucket-hybrid prediction, and direct residual regression.

We define the range proxy as $s_i=|h_1(z_i)|$.
The calibration procedure selects $h_1$ as the routing head and fixes the cut points before inference.
Six calibration-set cut points divide the proxy into seven range-proxy buckets indexed by $b$.
These routing buckets are distinct from the internal distance bins used by the first three heads.
Within bucket $b$, the four candidates are combined by bucketed convex distance stacking,
\begin{equation}
  r_i = \sum_{m=1}^{M} w_{b,m} c_{i,m},\quad w_{b,m}\ge 0,\quad \sum_m w_{b,m}=1,
  \label{eq:stack}
\end{equation}
Here, $M=4$. The weights in each range-proxy bucket are optimized by sequential least squares programming (SLSQP) to minimize the relative-error objective in Eq.~\ref{eq:rel}.
The stacked distance $r_i$ and candidate spread $\sigma_i=\operatorname{std}(c_{i,1},\ldots,c_{i,4})$ form the range statistics passed to the calibration module.

\begin{figure}[t]
  \centering
  \includegraphics[width=\linewidth]{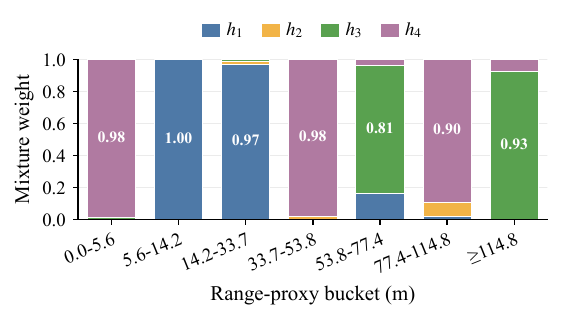}
  \caption{Mixture weights of the four distance heads across the seven range-proxy buckets.}
  \label{fig:weights}
\end{figure}


\subsection{Protocol-Specific Calibration Module}

The protocol-specific calibration module converts the range statistics into the final distance command.
The stacked estimate can retain bucket-dependent bias, which motivates a final regime-specific correction.
Within range-proxy bucket $b$, calibration-set quantiles assign $\sigma_i$ to one of three intervals.
This interval and the sign of $r_i$ jointly determine regime $q$.
Each combination $(b,q)$ selects an affine correction followed by 2.5\,m command-grid alignment,
\begin{equation}
  \hat d_i = Q_{2.5}\!\left(a_{b,q} r_i + c_{b,q}\right),
  \label{eq:snap}
\end{equation}
where $Q_{2.5}$ rounds its input to the nearest multiple of 2.5\,m.
This alignment is specific to the \task{} output protocol and should not be interpreted as the physical resolution of a UAV controller.

Distance recovery and heading calibration use separate branches in the pipeline.
The frozen heading predictor in Figure~\ref{fig:pipeline} supplies $\phi_i$, obtained from a fixed circular ensemble of per-pair heading estimates.
The released implementation stores this value in \texttt{heading\_pred}.
The fixed heading transform then applies the affine adjustment, 20-degree alignment, and angular wrapping shown in Eq.~\ref{eq:heading},
\begin{equation}
  \hat\theta_i =
  \operatorname{wrap}_{180}\!\left(
  20\cdot \operatorname{round}\frac{1.014\phi_i+1.2}{20}
  \right).
  \label{eq:heading}
\end{equation}
If the source and target image identifiers match, both outputs are set to zero. Otherwise, the system returns $(\hat\theta_i,\hat d_i)$ as the final command.
\section{Experiments}

Evaluation uses the public UAVM 2026 \task{} benchmark~\cite{li2026lastmeterprecisionnavigationuavs,deuser2026UVA}, which derives from University-1652 and contains more than 4.8 million image pairs across 72 scenes.
All model weights and calibration parameters were fixed before hidden-test inference.
The distance heads and protocol-specific calibration parameters were fitted on 367,416 image pairs.
A scene-disjoint set of 14,856 pairs was used for diagnostics, while 2,773,116 hidden pairs were reserved for official online evaluation.
We report relative distance error, relative angle error for heading estimation, and their mean as the final error; lower values indicate better performance.
The six range-proxy cut points, at 5.6, 14.2, 33.7, 53.8, 77.4, and 114.8\,m, were determined only from the calibration set.
Because the diagnostic set contains disjoint scenes, its results measure transfer to scenes not used for calibration.

\begin{table}[H]
  \caption{\task{} results. The \method{} row reports the archived official submission; the other rows provide reference comparisons. Bold indicates the best or tied-best value per metric.}
  \label{tab:main}
  \centering
  \setlength{\tabcolsep}{2.4pt}
  \begin{tabular}{@{}lccc@{}}
    \toprule
    Method & Final $\downarrow$ & Dist. rel. $\downarrow$ & Angle rel. $\downarrow$ \\
    \midrule
    SuperGlue reference~\cite{sarlin2020superglue} & 0.634000 & 0.988000 & 0.280000 \\
    Raw MASt3R, no calibration & 0.528784 & 0.979276 & 0.078291 \\
    Global scale calibration & 0.003424 & 0.003499 & \textbf{0.003350} \\
    \textbf{\method{}}  & \textbf{\score{}}  & \textbf{\distonline{}} & \textbf{\angleonline{}} \\
    \bottomrule
  \end{tabular}
\end{table}

Table~\ref{tab:main} summarizes performance on \task{}.
\method{} achieved a final error of \score{}, reducing the distance term relative to global scale calibration while leaving the angle term unchanged.
The large reduction from raw MASt3R to global calibration indicates that the frozen geometry contains a recoverable metric-scale signal.

\begin{table}[H]
  \caption{Ablation results on the scene-disjoint evaluation set. Checkmarks in the Calib. column denote protocol-specific calibration. Bold indicates the best or tied-best value per metric.}
  \label{tab:ablation}
  \centering
  \small
  \setlength{\tabcolsep}{2pt}
  \begin{tabular}{@{}llccc@{}}
    \toprule
    Variant & Calib. & Final $\downarrow$ & Dist. rel. $\downarrow$ & Angle rel. $\downarrow$ \\
    \midrule
    Global scale calibration  & \xmark & 0.005758 & 0.006656 & \textbf{0.004861} \\
    Global scale calibration  & \cmark & 0.004447 & 0.004033 & \textbf{0.004861} \\
    Bucketed convex stacking  & \xmark & 0.006403 & 0.007944 & \textbf{0.004861} \\
    \textbf{Full \method{}}  & \textbf{\cmark} & \textbf{0.003733} & \textbf{0.002605} & \textbf{0.004861} \\
    \midrule
    \quad Best single head    & \cmark & 0.006723 & 0.008585 & \textbf{0.004861} \\
    \quad Raw heading predictor & \cmark & 0.004579 & \textbf{0.002605} & 0.006553 \\
    \bottomrule
  \end{tabular}
\end{table}

Table~\ref{tab:ablation} isolates the contributions of range-aware distance estimation and protocol-specific calibration.
With protocol-specific calibration enabled, full \method{} reduced relative distance error from 0.004033 to 0.002605 compared with global scale calibration.
Bucketed convex stacking alone was insufficient: without protocol-specific calibration, its distance error was 0.007944, compared with 0.006656 for global scale calibration.
These results indicate that the gain depends on combining range-aware stacking with the final regime-specific correction.
Replacing the four-head pool with the best single head increased distance error to 0.008585.
Using the raw heading predictor left distance error unchanged but increased angle error, consistent with the separate distance and heading branches in the proposed pipeline.

\begin{figure}[H]
  \centering
  \includegraphics[width=\linewidth]{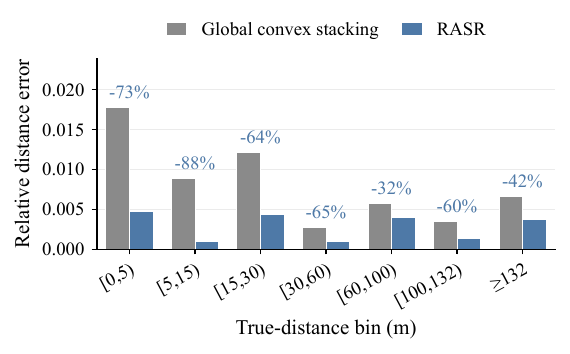}
  \caption{Relative distance error across true-distance bins for full \method{} and a global convex combination of the four candidate heads. Percentages indicate the relative reduction achieved by \method{}; true distance is used only for diagnostic binning.}
  \label{fig:range_calib}
\end{figure}

Compared with global convex stacking, full \method{} reduced relative distance error by 32\%--88\% across all true-distance bins.
These bins are used only for diagnosis, while inference routes each image pair using the predicted range proxy.
At inference, all reported variants process image pairs independently with fixed parameters, without batch sorting, cross-pair neighbor search, or revision of previous samples.

\section{Conclusion}

In this paper, we presented \method{} for estimating metric distance and heading between current and goal views in UAV navigation under GNSS denial. \method{} complements global scale calibration with range-aware residual correction and separates the scale-recovery core from a protocol-specific calibration module in a modular framework. On the scene-disjoint evaluation set, \method{} achieved lower relative distance error than global scale calibration under the same protocol-specific calibration. It also achieved a final error of \score{} on the official \task{} evaluation. These results show that frozen pairwise geometry can provide useful metric cues for per-pair navigation commands once its scale is calibrated and its range-dependent residuals are corrected.

\clearpage
{\small
\bibliographystyle{ACM-Reference-Format-citation-order}
\bibliography{refs}

@inproceedings{zhu2022transgeo,
  title     = {{TransGeo}: Transformer Is All You Need for Cross-View Image Geo-Localization},
  author    = {Zhu, Sijie and Shah, Mubarak and Chen, Chen},
  booktitle = {Proceedings of the IEEE/CVF Conference on Computer Vision and Pattern Recognition},
  pages     = {1162--1171},
  year      = {2022},
  publisher = {IEEE/CVF},
  address   = {New Orleans, LA, USA}
}

@article{zhang2023geodtr,
  title     = {Cross-View Geo-Localization via Learning Disentangled Geometric Layout Correspondence},
  author    = {Zhang, Xiaohan and Li, Xingyu and Sultani, Waqas and Zhou, Yi and Wshah, Safwan},
  journal   = {Proceedings of the AAAI Conference on Artificial Intelligence},
  volume    = {37},
  number    = {3},
  pages     = {3480--3488},
  year      = {2023}
}

@inproceedings{deuser2023sample4geo,
  title     = {{Sample4Geo}: Hard Negative Sampling for Cross-View Geo-Localisation},
  author    = {Deuser, Fabian and Habel, Konrad and Oswald, Norbert},
  booktitle = {Proceedings of the IEEE/CVF International Conference on Computer Vision},
  pages     = {16801--16810},
  year      = {2023},
  publisher = {IEEE/CVF},
  address   = {Paris, France}
}

@inproceedings{hu2022beyondgeoloc,
  title     = {Beyond Geo-localization: Fine-grained Orientation of Street-view Images by Cross-view Matching with Satellite Imagery},
  author    = {Hu, Wenmiao and Zhang, Yichen and Liang, Yuxuan and Yin, Yifang and Georgescu, Andrei and Tran, An and Kruppa, Hannes and Ng, See-Kiong and Zimmermann, Roger},
  booktitle = {Proceedings of the 30th ACM International Conference on Multimedia},
  pages     = {6155--6164},
  year      = {2022},
  publisher = {Association for Computing Machinery},
  address   = {New York, NY, USA},
  location  = {Lisboa, Portugal},
  doi       = {10.1145/3503161.3548102}
}

@inproceedings{mi2024congeo,
  title     = {{ConGeo}: Robust Cross-View Geo-Localization Across Ground View Variations},
  author    = {Mi, Li and Xu, Chang and Castillo-Navarro, Javiera and Montariol, Syrielle and Yang, Wen and Bosselut, Antoine and Tuia, Devis},
  booktitle = {Computer Vision -- ECCV 2024},
  pages     = {214--230},
  year      = {2024},
  publisher = {Springer Nature Switzerland},
  address   = {Cham, Switzerland}
}

@inproceedings{ye2025whereami,
  title     = {Where am I? Cross-View Geo-Localization with Natural Language Descriptions},
  author    = {Ye, Junyan and Lin, Honglin and Ou, Leyan and Chen, Dairong and Wang, Zihao and Zhu, Qi and He, Conghui and Li, Weijia},
  booktitle = {Proceedings of the IEEE/CVF International Conference on Computer Vision},
  pages     = {5890--5900},
  year      = {2025},
  publisher = {IEEE/CVF},
  address   = {Honolulu, HI, USA}
}

@inproceedings{zheng2020university1652,
  title     = {University-1652: A Multi-view Multi-source Benchmark for Drone-based Geo-localization},
  author    = {Zheng, Zhedong and Wei, Yunchao and Yang, Yi},
  booktitle = {Proceedings of the 28th ACM International Conference on Multimedia},
  pages     = {1395--1403},
  year      = {2020},
  publisher = {Association for Computing Machinery},
  address   = {New York, NY, USA}
}

@article{dai2022fsra,
  title   = {A Transformer-Based Feature Segmentation and Region Alignment Method for {UAV}-View Geo-Localization},
  author  = {Dai, Ming and Hu, Jianhong and Zhuang, Jiedong and Zheng, Enhui},
  journal = {IEEE Transactions on Circuits and Systems for Video Technology},
  volume  = {32},
  number  = {7},
  pages   = {4376--4389},
  year    = {2022}
}

@article{ji2025game4loc,
  title     = {{Game4Loc}: A {UAV} Geo-Localization Benchmark from Game Data},
  author    = {Ji, Yuxiang and He, Boyong and Tan, Zhuoyue and Wu, Liaoni},
  journal   = {Proceedings of the AAAI Conference on Artificial Intelligence},
  volume    = {39},
  number    = {4},
  pages     = {3913--3921},
  year      = {2025}
}

@article{dai2024uavselfpositioning,
  title   = {Vision-Based {UAV} Self-Positioning in Low-Altitude Urban Environments},
  author  = {Dai, Ming and Zheng, Enhui and Feng, Zhenhua and Lei, Qi and Zhuang, Jiedong and Yang, Wankou},
  journal = {IEEE Transactions on Image Processing},
  volume  = {33},
  pages   = {493--508},
  year    = {2024}
}

@article{liang2026synergetic,
  title   = {Synergetic Empowerment: Wireless Communications Meets Embodied Intelligence},
  author  = {Liang, Hongtao and Diao, Yihe and Wu, Yuhang and Zhou, Fuhui and Wu, Qihui},
  journal = {IEEE Communications Magazine},
  volume  = {64},
  pages   = {1--8},
  year    = {2026},
  doi     = {10.1109/MCOM.001.2500569}
}

@inproceedings{wang2024dust3r,
  title     = {{DUSt3R}: Geometric 3D Vision Made Easy},
  author    = {Wang, Shuzhe and Leroy, Vincent and Cabon, Yohann and Chidlovskii, Boris and Revaud, Jerome},
  booktitle = {Proceedings of the IEEE/CVF Conference on Computer Vision and Pattern Recognition},
  pages     = {20697--20709},
  year      = {2024},
  publisher = {IEEE/CVF},
  address   = {Seattle, WA, USA}
}

@inproceedings{leroy2024mast3r,
  title     = {Grounding Image Matching in 3D with {MASt3R}},
  author    = {Leroy, Vincent and Cabon, Yohann and Revaud, Jerome},
  booktitle = {Computer Vision -- ECCV 2024},
  pages     = {71--91},
  year      = {2024},
  publisher = {Springer Nature Switzerland},
  address   = {Cham, Switzerland}
}

@inproceedings{wang2025vggt,
  title     = {{VGGT}: Visual Geometry Grounded Transformer},
  author    = {Wang, Jianyuan and Chen, Minghao and Karaev, Nikita and Vedaldi, Andrea and Rupprecht, Christian and Novotny, David},
  booktitle = {Proceedings of the IEEE/CVF Conference on Computer Vision and Pattern Recognition},
  pages     = {5294--5306},
  year      = {2025},
  publisher = {IEEE/CVF},
  address   = {Nashville, TN, USA}
}

@inproceedings{yang2025fast3r,
  title     = {{Fast3R}: Towards 3D Reconstruction of 1000+ Images in One Forward Pass},
  author    = {Yang, Jianing and Sax, Alexander and Liang, Kevin J. and Henaff, Mikael and Tang, Hao and Cao, Ang and Chai, Joyce and Meier, Franziska and Feiszli, Matt},
  booktitle = {Proceedings of the IEEE/CVF Conference on Computer Vision and Pattern Recognition},
  pages     = {21924--21935},
  year      = {2025},
  publisher = {IEEE/CVF},
  address   = {Nashville, TN, USA}
}

@misc{zhang2025monst3r,
  title     = {{MonST3R}: A Simple Approach for Estimating Geometry in the Presence of Motion},
  author    = {Zhang, Junyi and Herrmann, Charles and Hur, Junhwa and Jampani, Varun and Darrell, Trevor and Cole, Forrester and Sun, Deqing and Yang, Ming-Hsuan},
  year      = {2025},
  howpublished = {International Conference on Learning Representations},
  eprint    = {2410.03825},
  archivePrefix = {arXiv},
  primaryClass = {cs.CV}
}

@inproceedings{wang2025cut3r,
  title     = {Continuous 3D Perception Model with Persistent State},
  author    = {Wang, Qianqian and Zhang, Yifei and Holynski, Aleksander and Efros, Alexei A. and Kanazawa, Angjoo},
  booktitle = {Proceedings of the IEEE/CVF Conference on Computer Vision and Pattern Recognition},
  pages     = {10510--10522},
  year      = {2025},
  publisher = {IEEE/CVF},
  address   = {Nashville, TN, USA}
}

@inproceedings{dong2025reloc3r,
  title     = {{Reloc3r}: Large-Scale Training of Relative Camera Pose Regression for Generalizable, Fast, and Accurate Visual Localization},
  author    = {Dong, Siyan and Wang, Shuzhe and Liu, Shaohui and Cai, Lulu and Fan, Qingnan and Kannala, Juho and Yang, Yanchao},
  booktitle = {Proceedings of the IEEE/CVF Conference on Computer Vision and Pattern Recognition},
  pages     = {16739--16752},
  year      = {2025},
  publisher = {IEEE/CVF},
  address   = {Nashville, TN, USA}
}

@inproceedings{yin2023metric3d,
  title     = {{Metric3D}: Towards Zero-shot Metric 3D Prediction from A Single Image},
  author    = {Yin, Wei and Zhang, Chi and Chen, Hao and Cai, Zhipeng and Yu, Gang and Wang, Kaixuan and Chen, Xiaozhi and Shen, Chunhua},
  booktitle = {Proceedings of the IEEE/CVF International Conference on Computer Vision},
  pages     = {9043--9053},
  year      = {2023},
  publisher = {IEEE/CVF},
  address   = {Paris, France}
}

@article{hu2024metric3dv2,
  title   = {{Metric3D v2}: A Versatile Monocular Geometric Foundation Model for Zero-shot Metric Depth and Surface Normal Estimation},
  author  = {Hu, Mu and Yin, Wei and Zhang, Chi and Cai, Zhipeng and Long, Xiaoxiao and Chen, Hao and Wang, Kaixuan and Yu, Gang and Shen, Chunhua and Shen, Shaojie},
  journal = {IEEE Transactions on Pattern Analysis and Machine Intelligence},
  volume  = {46},
  number  = {12},
  pages   = {10579--10596},
  year    = {2024}
}

@inproceedings{piccinelli2024unidepth,
  title     = {{UniDepth}: Universal Monocular Metric Depth Estimation},
  author    = {Piccinelli, Luigi and Yang, Yung-Hsu and Sakaridis, Christos and Segu, Mattia and Li, Siyuan and Van Gool, Luc and Yu, Fisher},
  booktitle = {Proceedings of the IEEE/CVF Conference on Computer Vision and Pattern Recognition},
  pages     = {10106--10116},
  year      = {2024},
  publisher = {IEEE/CVF},
  address   = {Seattle, WA, USA}
}

@inproceedings{yang2024depthanything,
  title     = {Depth Anything: Unleashing the Power of Large-Scale Unlabeled Data},
  author    = {Yang, Lihe and Kang, Bingyi and Huang, Zilong and Xu, Xiaogang and Feng, Jiashi and Zhao, Hengshuang},
  booktitle = {Proceedings of the IEEE/CVF Conference on Computer Vision and Pattern Recognition},
  pages     = {10371--10381},
  year      = {2024},
  publisher = {IEEE/CVF},
  address   = {Seattle, WA, USA}
}

@inproceedings{yang2024depthanythingv2,
  title   = {Depth Anything V2},
  author  = {Yang, Lihe and Kang, Bingyi and Huang, Zilong and Zhao, Zhen and Xu, Xiaogang and Feng, Jiashi and Zhao, Hengshuang},
  booktitle = {Advances in Neural Information Processing Systems},
  volume  = {37},
  pages   = {21875--21911},
  year    = {2024},
  publisher = {Curran Associates, Inc.},
  address = {Vancouver, BC, Canada}
}

@inproceedings{ke2024marigold,
  title     = {Repurposing Diffusion-Based Image Generators for Monocular Depth Estimation},
  author    = {Ke, Bingxin and Obukhov, Anton and Huang, Shengyu and Metzger, Nando and Daudt, Rodrigo Caye and Schindler, Konrad},
  booktitle = {Proceedings of the IEEE/CVF Conference on Computer Vision and Pattern Recognition},
  pages     = {9492--9502},
  year      = {2024},
  publisher = {IEEE/CVF},
  address   = {Seattle, WA, USA}
}

@inproceedings{shah2023gnm,
  title     = {{GNM}: A General Navigation Model to Drive Any Robot},
  author    = {Shah, Dhruv and Sridhar, Ajay and Bhorkar, Arjun and Hirose, Noriaki and Levine, Sergey},
  booktitle = {Proceedings of the IEEE International Conference on Robotics and Automation},
  pages     = {7226--7233},
  year      = {2023},
  publisher = {IEEE},
  address   = {London, United Kingdom}
}

@inproceedings{shah2023vint,
  title     = {{ViNT}: A Foundation Model for Visual Navigation},
  author    = {Shah, Dhruv and Sridhar, Ajay and Dashora, Nitish and Stachowicz, Kyle and Black, Kevin and Hirose, Noriaki and Levine, Sergey},
  booktitle = {Proceedings of the 7th Conference on Robot Learning},
  series    = {Proceedings of Machine Learning Research},
  volume    = {229},
  pages     = {711--733},
  year      = {2023},
  publisher = {PMLR},
  address   = {Atlanta, GA, USA}
}

@inproceedings{sridhar2024nomad,
  title     = {{NoMaD}: Goal Masked Diffusion Policies for Navigation and Exploration},
  author    = {Sridhar, Ajay and Shah, Dhruv and Glossop, Catherine and Levine, Sergey},
  booktitle = {Proceedings of the IEEE International Conference on Robotics and Automation},
  pages     = {63--70},
  year      = {2024},
  publisher = {IEEE},
  address   = {Yokohama, Japan}
}

@misc{deng2026anyimagenav,
  title         = {{AnyImageNav}: Any-View Geometry for Precise Last-Meter Image-Goal Navigation},
  author        = {Deng, Yijie and Yuan, Shuaihang and Fang, Yi},
  year          = {2026},
  eprint        = {2604.05351},
  archivePrefix = {arXiv},
  primaryClass  = {cs.RO}
}

@inproceedings{sarlin2020superglue,
  title     = {{SuperGlue}: Learning Feature Matching with Graph Neural Networks},
  author    = {Sarlin, Paul-Edouard and DeTone, Daniel and Malisiewicz, Tomasz and Rabinovich, Andrew},
  booktitle = {Proceedings of the IEEE/CVF Conference on Computer Vision and Pattern Recognition},
  pages     = {4938--4947},
  year      = {2020},
  publisher = {IEEE/CVF},
  address   = {Seattle, WA, USA}
}

@inproceedings{lindenberger2023lightglue,
  title     = {{LightGlue}: Local Feature Matching at Light Speed},
  author    = {Lindenberger, Philipp and Sarlin, Paul-Edouard and Pollefeys, Marc},
  booktitle = {Proceedings of the IEEE/CVF International Conference on Computer Vision},
  pages     = {17581--17592},
  year      = {2023},
  publisher = {IEEE/CVF},
  address   = {Paris, France}
}

@misc{li2026lastmeterprecisionnavigationuavs,
  title         = {Last-Meter Precision Navigation for UAVs: A Diffusion-Refined Aerial Visual Servoing Approach},
  author        = {Li, Yaxuan and Zeng, Jiarui and Huang, Shaofei and Zheng, Zhedong},
  year          = {2026},
  eprint        = {2607.04352},
  archivePrefix = {arXiv},
  primaryClass  = {cs.CV},
  url           = {https://arxiv.org/abs/2607.04352}
}

@inproceedings{deuser2026UVA,
  title     = {The 4th Workshop on UAVs in Multimedia: Capturing the World from a New Perspective},
  author    = {Deuser, Fabian and Li, Yaxuan and Wang, Tingyu and Shi, Yujiao and B{\"o}{\ss}end{\"o}rfer, Anna and Huang, Shaofei and Pan, Xiao and Zheng, Zhedong and Zimmermann, Roger},
  booktitle = {Proceedings of the 34th ACM International Conference on Multimedia Workshop},
  year      = {2026},
  publisher = {Association for Computing Machinery},
  address   = {New York, NY, USA},
  numpages  = {3}
}
}

\end{document}